\documentclass[english, 10pt, twocolumn, twoside]{IEEEtran}
\usepackage{threeparttable}
\usepackage{mathrsfs}

\usepackage{cite}

\ifCLASSINFOpdf
\else
\fi
\usepackage{graphics}

\usepackage{subfigure}
\usepackage[cmex10]{amsmath}
\usepackage{amsfonts}
\usepackage{cases}
\usepackage{epsfig}
\usepackage{url}

\usepackage{algorithm}
\usepackage{algorithmic}

\usepackage{array}
\usepackage{amsmath}

\usepackage{multirow}
\usepackage{bigstrut}

\usepackage{bm}
\usepackage{amsmath}
\usepackage{amssymb}
\usepackage{mathrsfs}
\usepackage{color}

\usepackage{booktabs}
\usepackage{multirow}
\usepackage{amsmath}
\usepackage[bottom]{footmisc}
\usepackage{bbm}

\newcommand{\comment}[1]{\textcolor[rgb]{1.0,0.00,0.00}{#1}}

\usepackage{graphicx}
\usepackage{amsfonts}       
\usepackage{nicefrac}       
\usepackage{microtype}      
\usepackage{xcolor}         

\usepackage{amsthm}
\usepackage{threeparttable}
\usepackage{bigstrut}
\usepackage{float}

\usepackage{amsthm,amsmath,amssymb}
\usepackage{mathrsfs}

\usepackage[algo2e]{algorithm2e}

\newcommand{\cskip}{c_{\text{skip}}}
\newcommand{\cin}{c_{\text{in}}}
\newcommand{\cout}{c_{\text{out}}}
\newcommand{\cnoise}{c_{\text{noise}}}

\usepackage{wrapfig} 
\usepackage{caption} 

\begin{document}

%
\title{Zero-Shot Video Restoration and Enhancement with Assistance of Video Diffusion Models}
%
%
%

\author{Cong Cao, Huanjing Yue, \textit{Senior Member, IEEE}, Shangbin Xie, Xin Liu, Jingyu Yang, \textit{Senior Member, IEEE}
\thanks{This work was supported in part by the National Natural Science Foundation of China under Grant 62472308 and Grant 62231018. C. Cao, H. Yue (corresponding author), S. Xie and J. Yang are with the School of Electrical and Information Engineering, Tianjin University, Tianjin, China, 300072. X. Liu is with the School of Psychology, Shanghai Jiao Tong University, China.}}

%
%

\markboth{}%
{Cao \MakeLowercase{\textit{et al.}}: Zero-Shot Video Restoration and Enhancement with Assistance of Video Diffusion Models}
%



\maketitle

\begin{abstract}
Although diffusion-based zero-shot image restoration and enhancement methods have achieved great success, applying them to video restoration or enhancement will lead to severe temporal flickering. In this paper, we propose to 
systematically utilize both homologous and heterogeneous video diffusion models to assist the image-based method in maintaining more temporal consistency for zero-shot video restoration and enhancement. We propose homologous latents fusion, heterogenous latents fusion, and a CoQ fusion ratio strategy to utilize both homologous and heterogenous text-to-video diffusion models to complement the image method. Moreover, we propose temporal-strengthening post-processing to utilize the image-to-video diffusion model to further improve temporal consistency. Our method is training-free and can be applied to any diffusion-based image restoration and enhancement methods. Experimental results demonstrate the superiority of the proposed method.
\end{abstract}

\begin{IEEEkeywords}
Zero-shot, Video Restoration, Video Enhancement, Text-to-video Diffusion Model, Image-to-video Diffusion Model.
\end{IEEEkeywords}

%
\IEEEpeerreviewmaketitle

\section{Introduction}

Image and video restoration (IVR) has entered a new era with the rise of diffusion models.
By leveraging powerful generative priors, zero-shot restoration methods \cite{rout2024solving, chung2023prompt, he2023iterative} have achieved remarkable progress, offering a versatile alternative to supervised/unsupervised learning without the need for training data. 
These advancements have enabled high-fidelity image restoration (IR) across various degradation types through latent space manipulation and guided sampling processes. 

Despite these successes, a critical gap remains between zero-shot image restoration and its video counterpart. While text-to-image (T2I) diffusion models excel at reconstructing spatial details, they lack the temporal knowledge required for video sequences. Directly applying image-based restoration to videos leads to severe temporal flickering, as each frame is processed in isolation. Recent attempts to bridge this gap primarily rely on inserting training-free temporal modules (e.g., short-long-range temporal attention and noise-sharing \cite{cao2025zero} or flow-guided token merging \cite{yeh2024diffir2vr}) into existing IR frameworks. 
However, these methods are fundamentally constrained by their "training-free" nature and cannot truly understand complex motion dynamics.

The emergence of video diffusion models (T2V/I2V) presents a promising frontier, as they possess robust temporal priors learned from vast video datasets. Yet, effectively harvesting these priors for zero-shot video restoration introduces three challenges. First, architectural heterogeneity: Most state-of-the-art (SOTA) T2V models utilize specialized 3D VAEs that are structurally incompatible with standard IR frameworks. This heterogeneity prevents direct latent fusion, restricting most current methods to a narrow subset of ``homologous" models. Second, temporal-fidelity trade-off: Determining the optimal fusion ratio between restoration latents and video priors is a non-trivial challenge. A conservative ratio fails to suppress flickering, while an aggressive one overpowers the restoration model, leading to blurred details or structural distortion. Third, inherent flicker residuals: Even with latent fusion, the high variance in the base IR model's outputs can still result in residual flickering that latent-level constraints alone cannot fully resolve.

To address these challenges, we introduce ZVRV, a novel framework for Zero-shot Video Restoration with Video Diffusion Assistance. Our approach transcends the limitations of training-free modules by directly injecting pre-trained temporal knowledge into the restoration process. Specifically, we propose a Heterogeneous Latent Fusion mechanism that bypasses VAE restrictions, enabling our framework to leverage any SOTA video diffusion model. To solve the fusion ratio dilemma, we introduce a Chain-of-Quality (CoQ) based Fusion Strategy. Inspired by the reasoning capabilities of Large Language Models, we treat the selection of fusion ratios as a sequential decision task. We utilize the video quality metric as a test-time verifier to sample and select the optimal ratio at each timestep, ensuring a precise balance between temporal stability and spatial fidelity. Finally, we incorporate a temporal-strengthening post-processing stage using I2V models to further refine the results. When T2V models utilize a different sampling method from the image IR model (such as DDIM sampling and flow-matching sampling) or when the image IR model is a one-step diffusion model, latent fusion cannot be utilized. However, our temporal-strengthening post-processing can still be applied to enhance temporal consistency. 

Video diffusion models have recently been explored for zero-shot video restoration \cite{spagnoletti2026lvtino,varanka2026zero,kwon2026accelerating}. These methods, however, are built exclusively upon video generation models and do not incorporate image restoration/enhancement models into the restoration pipeline. As a consequence, the degradation information that image restoration/enhancement models are able to learn remains unexploited, and the resulting frameworks are therefore limited in their ability to handle complex degradations. In contrast to these approaches, the proposed method employs video generation models as an auxiliary component to assist image restoration/enhancement models. This design enables the joint exploitation of two complementary sources of information: the temporal prior provided by the video generation model and the degradation information captured by the image restoration/enhancement model, thereby improving performance under complex degradation conditions.

Our contributions are summarized as follows. 

\begin{itemize}\item[$\bullet$] A Universal Zero-shot Framework. We propose ZVRV, the first framework that systematically utilizes both homologous and heterogeneous video diffusion models to assist zero-shot video restoration.

\item[$\bullet$] CoQ Latent Fusion and Temporal-strengthening Post-processing. We introduce a novel CoQ latent fusion strategy utilizing a Process Reward Model as a test-time verifier, enabling the adaptive selection of optimal fusion ratios across diffusion timesteps. Our temporal-strengthening post-processing further overcomes the limitations of the IR model's sampling method and steps, enabling it to be applied to any IR model.

\item[$\bullet$] Extensive experiments demonstrate that our method effectively overcomes the limitations of training-free temporal modules, achieving state-of-the-art temporal consistency in zero-shot video restoration.\end{itemize}

\section{Related Work}

\subsection{Zero-Shot IR with Latent Diffusion Models}


The success of diffusion generative models enlightened zero-shot image restoration methods, which can be further devided in pixel-space zero-shot IR and latent-space zero-shot IR. Pixel-space zero-shot IR \cite{song2019generative,lugmayr2022repaint,choi2021ilvr,kawar2022denoising,wang2022zero,chung2022diffusion,fei2023generative} utilizes pretrained unconditional image diffusion models working on pixel-space, like \cite{dhariwal2021diffusion}. Latent-space zero-shot IR \cite{rout2024solving,chung2023prompt,he2023iterative,rout2023beyond} utilizes pretrained text-to-image latent diffusion models, like Stable Diffusion \cite{rombach2022high}. PSLD \cite{rout2024solving} is the first framework to solve zero-shot image restoration with text-to-image latent diffusion model. \cite{chung2023prompt} proposes a prompt tuning method to jointly optimizes the text embedding in the sampling. \cite{he2023iterative} uses the historical gradient information to guide the sampling. But all these methods are designed for image recovery problems, there exists severe temporal flickering when apply them to degraded videos.

\subsection{Zero-Shot Video Editing}

Along with the development of powerful pre-trained text-to-image diffusion models like Stable Diffusion \cite{rombach2022high}, diffusion-based zero-shot video editing \cite{wu2023tune,zhao2023controlvideo,yang2023rerender} has gained increasing attention, which utilizes these off-the-shelf text-to-image diffusion model and mainly solve the temporal consistency problem. FateZero \cite{qi2023fatezero} follows Prompt-to-Prompt \cite{hertz2022prompt} and fuse the attention maps in the DDIM inversion process and generation process to preserve the motion and structure consistency. Text2Video-Zero \cite{khachatryan2023text2video} proposes cross-frame attention and motion dynamics to enrich the latent codes for better temporal consistency. FVDM \cite{lu2024fuse} proposes to fuse the latents between T2I and T2V latents for zero-shot video editing. And FVDM can be regarded as a kind of homologous latents fusion. Different from this work, we transfer the homologous latents fusion for video restoration and enhancement, and further propose the heterogenous latents fusion to get rid of the restrictions on T2V models. Through our homologous and heterogenous latents fusion, we can utilize any kind of SOTA T2V model to assist the image restoration/enhancement model in achieving temporal consistent video restoration/enhancement. In addition, we propose a novel CoQ Fusion Ratio Strategy to better control the fusion ratio. And we propose temporal-strengthening post-processing to utilize the I2V model to further improve the temporal consistency.

\subsection{Zero-Shot Video Restoration and Enhancement}

ZVRD \cite{cao2025zero} proposes the short-long-range temporal attention layer, temporal consistency guidance, spatial-temporal noise sharing, and an early stopping sampling strategy for zero-shot video restoration. DiffIR2VR \cite{yeh2024diffir2vr} proposes flow-guided video token merging to improve temporal consistency when applying diffusion-based image restoration models for video restoration. Although these methods design training-free temporal modules and insert them into image restoration/enhancement method for video restoration/enhancement, their temporal modules are also limited by training-free manner. Recently, more and more amazing video diffusion models have emerged. These video diffusion models have been trained on a large amount of video data to learn powerful temporal priors which can preserve the temporal consistency of results. We propose to utilize the temporal priors in these video diffusion models to overcome the limitations of previous training-free temporal modules. Although some recent works also have utilized video diffusion models for zero-shot video restoration \cite{spagnoletti2026lvtino,varanka2026zero,kwon2026accelerating}, they have employed only video generation models without leveraging image restoration/enhancement models, and thus cannot exploit the degradation information learned by some image restoration/enhancement models, making it difficult to apply them to complex degradations. Unlike them, we utilize video generation models to assist image restoration/enhancement models, enabling the joint exploitation of the temporal prior from the former and the degraded information from the latter.

\begin{figure*}
    \centering
    \includegraphics[width=1.0\linewidth]{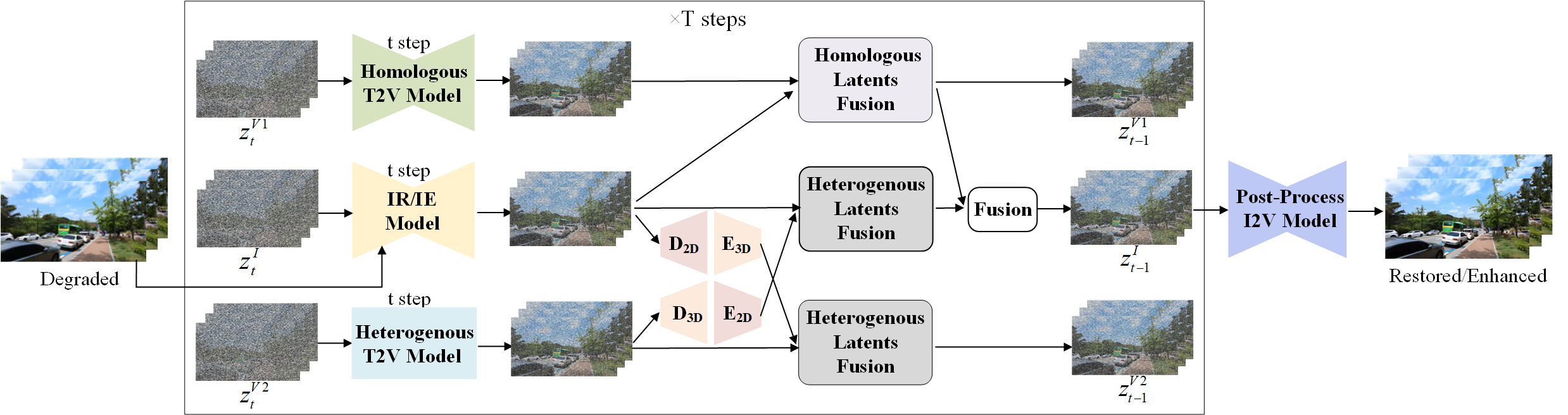}
    \caption{Overview of our framework.}
    \label{fig:framework}
\end{figure*}

\section{Background}

\subsection{Latent Diffusion Models}

Latent diffusion models (LDM) operate in the latent space with VAE autoencoder $\mathcal{E}$, $\mathcal{D}$. First, an encoder $\mathcal{E}$ compresses the input RGB image/video $x$ to a low-resolution latent $z=\mathcal{E}(x)$, the forward and reverse diffusion process work on the latent, the latent can be reconstructed back to image/video $ \mathcal{D}(z) \approx x $ by decoder $\mathcal{D}$. In the forward diffusion process, Gaussian noise is gradually added to $z_0$ to obtain $z_t$ through Markov transition with the transition probability
\begin{equation}
\centering
q(z_t | z_{t-1}) = \mathcal{N}(z_t; \sqrt{1-\beta_t}z_{t-1}, \beta_t \text{I})
\label{eq:ddpm_forward}
\end{equation}
where $\beta_t$ is the variance schedule for the timestep $t$. The backward process uses a trained U-Net $\varepsilon_{\theta}$ for denoising:
\begin{equation}
\centering
p_{\theta}(z_{t-1}|z_t) = \mathcal{N}(z_{t-1}; \mu_{\theta}(z_t, \tau, t),  \Sigma_{\theta}(z_t, \tau, t) )
\label{eq:ddpm_backward}
\end{equation}
where $\tau$ denotes the textual prompt. $\mu_{\theta}$ and $\Sigma_{\theta}$ are computed by $\varepsilon_{\theta}$.

\subsection{DDIM Sampling}

DDIM sampling \cite{song2020denoising} is employed to reverse diffusion process, which converts noisy latent $z_T$ to a clean latent $z_0$ in a sequence of timestep:
\begin{equation}
\label{eq: denoise}
    z_{t-1} = \sqrt{\alpha_{t-1}} \; \frac{z_t - \sqrt{1-\alpha_t}{\varepsilon_\theta}}{\sqrt{\alpha_t}}+ \sqrt{1-\alpha_{t-1}-\sigma_{t}^2}{\varepsilon_\theta} + \sigma_{t}\epsilon_{t}
\end{equation}
where $\alpha_{t}$ and $\sigma_{t}$ are parameters for noise scheduling \cite{song2020denoising}, $\epsilon_{t} \sim \mathcal{N}(0, 1)$. In practice, firstly $\boldsymbol{\hat{z}}_0$ is predicted from $\boldsymbol{z}_t$
\begin{equation}
\boldsymbol{\hat{z}}_{0} =  \frac{\boldsymbol{z}_{t}}{\sqrt{\bar{\alpha}_{t}}}-\frac{\sqrt{1-\bar{\alpha}_{t}} \varepsilon_{\theta}}{\sqrt{\bar{\alpha}_{t}}}
\label{z0}
\end{equation}
where $\bar{\alpha}_t=\prod_{i=1}^t \alpha_i$. Then $\boldsymbol{z}_{t-1}$ is sampled using both $\boldsymbol{\hat{z}}_0$ and $\boldsymbol{z}_t$
\begin{equation}
    z_{t-1} = \frac{\sqrt{\alpha_t}(1-\bar{\alpha}_{t-1})}{1 - \bar{\alpha}_t}z_t + \frac{\sqrt{\bar{\alpha}_{t-1}}\beta_t}{1 -\bar{\alpha}_t}\hat{z}_0 +  \sigma_{t}\epsilon_{t}
\end{equation}
where $\beta_t = 1-\alpha_t$.

\subsection{Latent Diffusion Models for Zero-Shot IR}

Linear inverse problem in image restoration (IR) can be formulated as
\begin{equation}
    y = Ax + n
\end{equation}
where $A$ is the linear degradation operator and $n$ is additive white Gaussian noise, the task is restoring the ground-truth image $x$ from the degraded image $y$. Following PSLD \cite{rout2024solving}, latent constraint is applied in the reverse diffusion process to preserve the content between generated result and degraded image. The constraint loss is formulated as
\begin{equation}
\begin{split}
\mathcal{L} &= \mathcal{L}_{rec} + \gamma_1\mathcal{L}_{reg} \\
\mathcal{L}_{rec} &= \|y - A(\mathcal{D}(\hat{z}_0))\|_2^2 \\
\mathcal{L}_{reg} &= \|\hat z_0 - \mathcal{E}({A^Ty + (I - A^TA) \mathcal{D}(\hat z_0))}\|_2^2
\end{split}
\label{loss}
\end{equation}
where $\mathcal{L}_{rec}$ directly constrains the content, $\mathcal{L}_{reg}$ penalizes latents that are not fixed-points of the composition of the decoder-function with the encoder-function, make sure that the generated sample remains on the manifold of real data.

\section{Method}

Given a degraded video with $N$ frames $\{I_i\}_{i=0}^N$, our goal is to restore/enhance it to a clean/normal-light video $\{I''_i\}_{i=0}^N$. Our method leverages the homologous and heterogenous T2V models to assist in image restoration, achieving better temporal consistency and visual quality. We propose the corresponding homologous and heterogenous latents fusion, and further propose a CoQ fusion ratio strategy to update the fusion ratio self-adaptively. The framework is illustrated in Fig. \ref{fig:framework}. Our method is training-free and can be applied to any LDM-based image restoration methods, including zero-shot IR and trained IR method.

\subsection{Homologous Latents Fusion}

In the early stage of T2V model development, temporal modules are incorporated into the 2D denoising UNet from T2I LDM (Stable Diffusion) to construct the 3D denoising UNet for T2V LDM model, and the 2D VAE Encoder and Decoder from T2I are maintained for the T2V model. The representative T2V models are ModelScopeT2V \cite{wang2023modelscope} and ZeroScope \cite{zeroscope}. For the Stable-Diffusion-based image restoration methods \cite{rout2024solving,lin2024diffbir}, these T2V models share the same 2D VAE as those methods. We called this kind of T2V model the homologous T2V model. The latents between image restoration method and homologous T2V model can be directly fused, we called them homologous latents. We proposed to fuse the homologous latents between IR method and homologous T2V model to improve the temporal consistency.

To be specific, the image restoration/enhancement (IR/IE) model and homologous T2V model share the same initial noisy latents $z_T$, then the IR/IE model and homologous T2V model are applied to predict noise and generate the noisy latents $z_t^I$ and $z_t^{V1}$ by DDIM sampling, respectively. Different from T2V model, The IR/IE method predict the noise frame-by-frame and the denoised latents are concatenated along the temporal dimension. Then the noisy latents $z_t^I$ and $z_t^{V1}$ are fused to generate the fused noisy latents $z_t^{F1}$
\begin{equation}
    z_t^{F1} = (1-\lambda_t^{F1})z_t^I + \lambda_t^{F1}z_t^{V1}
\end{equation}
where the hyper-parameter $\lambda_t^{F1}$ denoted the fusion weight. Then the IR/IE model and homologous T2V model share the fused noisy latents $z_t^{F1}$ and predict noise for the next denoising timestep. 

\subsection{Heterogenous Latents Fusion}

Along with the development of T2V model, 2D VAE is replaced by 3D VAE, and the 3D denoising UNet is replace by MM-DiTs. The representative T2V models are CogVideoX \cite{yang2024cogvideox} and HunyuanVideo \cite{kong2024hunyuanvideo}, which have better consistency with the prompt, better temporal consistency and aesthetic quality than the previous T2V model. These T2V models have the different VAE from Stable-Diffusion-based image restoration methods, their latents can not be directly fused. We called this kind of T2V model the heterogenous T2V model. We further propose the heterogenous latents fusion to utilize heterogenous T2V model to improve the temporal consistency and quality of video results.

Given the degraded video $\{I_i\}_{i=0}^N$, we encode it to latents $z_0^{V2}$ with the 3D VAE. Then add noise to the latents $z_0^{V2}$ to achieve initial noisy latents $z_T^{V2}$. Then we apply heterogenous T2V model to predict noise and generate the noisy latents $z_t^{V2}$ by DDIM sampling. The clean latents $\hat{z}_0^{V2}$ can be predicted from $z_t^{V2}$ by the Eq. \ref{z0}. We decode the clean latents $\hat{z}_0^{V2}$ to video $\{I^{V2}_i\}_{i=0}^N$ by 3D VAE decoder $\mathcal{D}_{3D}$. Then we encode the video $\{I^{V2}_i\}_{i=0}^N$ to latents $\hat{z}_0^{V2\rightarrow I}$ by 2D VAE encoder $\mathcal{E}_{2D}$ 
\begin{equation}
    \hat{z}_0^{V2\rightarrow I} = \mathcal{E}_{2D}(\mathcal{D}_{3D}(\hat{z}_0^{V2}))
\end{equation}
The clean latents $\hat{z}_0^{I}$ can also be predicted from $z_t^{I}$ by the Eq. \ref{z0}. Then the clean latents $\hat{z}_0^I$ and $\hat{z}_0^{V2\rightarrow I}$ are fused to generate the fused clean latents $z_0^{F2}$, 
\begin{equation}
\begin{split}
    \hat{z}_0^{F2} &= (1-\lambda_t^{F2})\hat{z}_0^I + \lambda_t^{F2}\hat{z}_0^{V2\rightarrow I} \\
\end{split}
\end{equation}
where the hyper-parameter $\lambda_t^{F2}$ are the fusion weights.
According to Eq. \ref{z0}, the corresponding fused noisy latents $z_t^{F2}$  can be achieved by
\begin{equation}
    z_t^{F2} = z_t^{I} + \sqrt{\bar{\alpha}_{t}^{I}}(\hat{z}_0^{F2}-\hat{z}_0^{I})
\end{equation}
where $\bar{\alpha}_{t}^{I}$ denotes the $\bar{\alpha}_{t}$ in the sampling of IR/IE model. $z_t^{F2}$ is fused with $z_t^{F1}$ to get the final fused latents for the IR/IE model
\begin{equation}
\begin{split}
    z_t^{F} &= (1-\lambda_t^{F})z_t^{F1} + \lambda_t^{F}z_t^{F2}
\end{split}
\end{equation}
where the hyper-parameter $\lambda_t^{F}$ are the fusion weights. In a similar way, we can convert the IR/IE model latents to the latent space of heterogenous T2V model latents to guide the sampling of heterogenous T2V model.  We decode the clean latents $\hat{z}_0^{I}$ to video $\{I^{I}_i\}_{i=0}^N$ by 2D VAE decoder $\mathcal{D}_{2D}$ and encode it to latents $\hat{z}_0^{I\rightarrow V2}$ by 3D VAE encoder $\mathcal{E}_{3D}$
\begin{equation}
    \hat{z}_0^{I\rightarrow V2} = \mathcal{E}_{3D}(\mathcal{D}_{2D}(\hat{z}_0^{I}))
\end{equation}
Then the latents $\hat{z}_0^{V2}$ and $\hat{z}_0^{I\rightarrow V2}$ are fused to generate the fused clean latents $\hat{z}_0^{FV2}$ for the heterogenous T2V model
\begin{equation}
    \hat{z}_0^{FV2} = \lambda_t^{F2}\hat{z}_0^{V2} + (1-\lambda_t^{F2})\hat{z}_0^{I\rightarrow V2}
\end{equation}
The corresponding fused noisy latents $z_t^{FV2}$ can be achieved by
\begin{equation}
    z_t^{FV2} = z_t^{V2} + \sqrt{\bar{\alpha}_{t}^{V2}}(\hat{z}_0^{FV2}-\hat{z}_0^{V2})
\end{equation}
where $\bar{\alpha}_{t}^{V2}$ denotes the $\bar{\alpha}_{t}$ in the sampling of heterogenous T2V model.
Although the heterogeneous T2V model has better temporal consistency and video quality, heterogenous latents fusion alone is not optimal due to information loss during VAE encoding and decoding. Combining homologous and heterogenous latents through effective strategy can further improve the performance.

\begin{figure}
    \centering
    \includegraphics[width=1.0\linewidth]{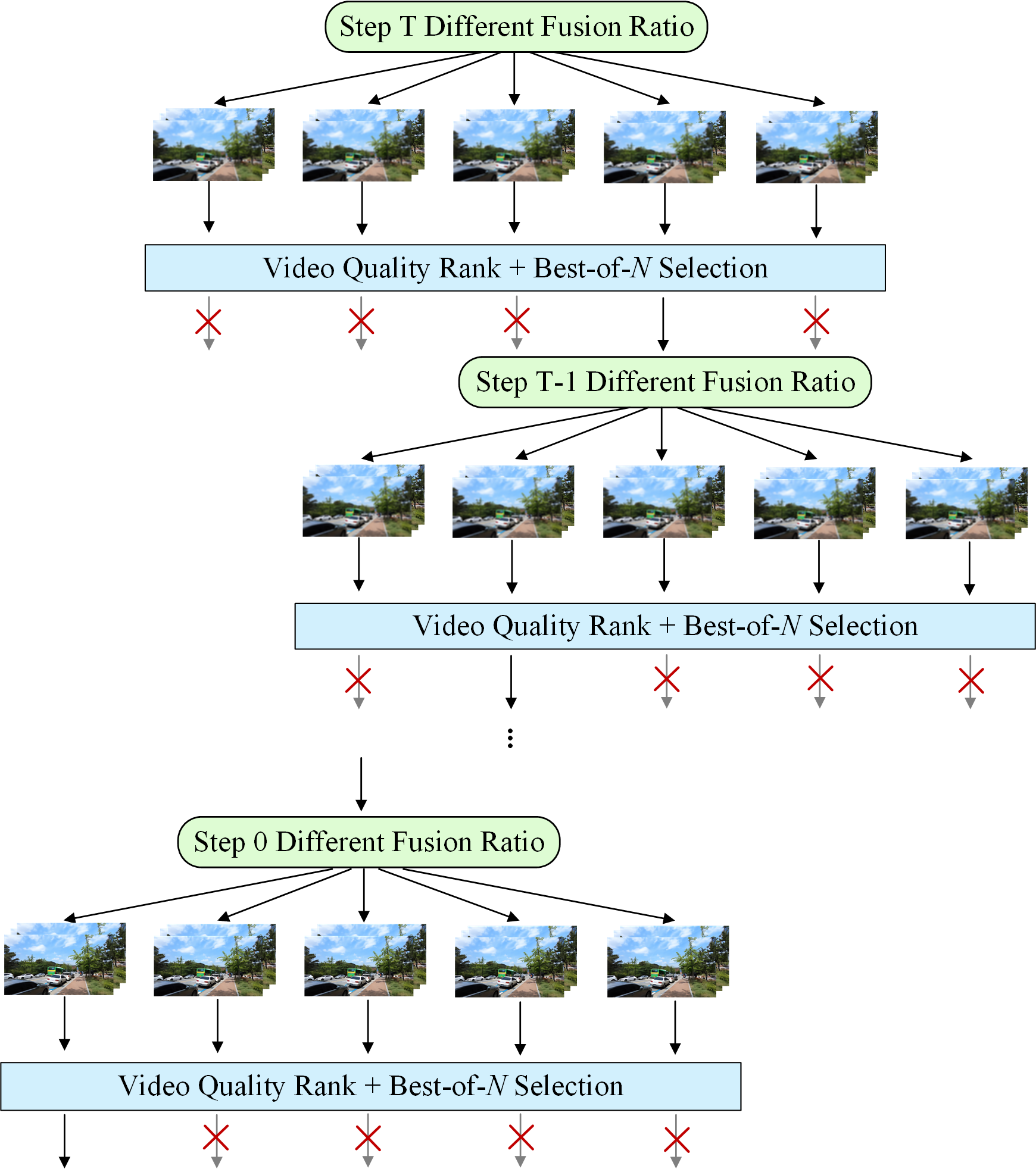}
    \caption{Our CoQ Fusion Ratio Strategy. We utilize video quality metrics as test-time verifiers within searching paths.}
    \label{fig:fusioratiostrategy}
\end{figure}

\subsection{CoQ Fusion Ratio Strategy}

How to set the fusion ratio $\lambda_t^{F1}$, $\lambda_t^{F2}$ and $\lambda_t^{F}$ at different timestep $t$ is a challenge problem. If $\lambda_t^{F1}$ and $\lambda_t^{F2}$ are too small, it would not work. Too large $\lambda_t^{F1}$ and $\lambda_t^{F2}$ will result in strong temporal consistent but blurry results or destroy the structure. Inspired by the development of Chain-of-Thought (COT) \cite{lightman2023let,ma2023let,snell2024scaling,guo2025can}, we propose Chain-of-Quality (CoQ) Fusion Ratio Strategy to solve this problem, which is shown in Fig. \ref{fig:fusioratiostrategy}. CoT reasoning has been widely adopted to address complex problems in language understanding and multimodal perception through sequential, step-by-step reasoning. We found that a similar sequential, step-by-step searching for the optimal fusion ratio during the sampling process can further improve the performance. Unlike CoT, which often employs reward models as test-time verifiers within its reasoning paths, we utilize video quality metric as test-time verifiers within searching paths. We adopt the best-of-N selection for test-time verifiers.


Take $\lambda_t^{F1}$ as an example, we uniformly sample $M+1$ values of $\lambda_T^{F1}$ from [$\lambda_t^{c}-r$, $\lambda_t^{c}+r$] at timestep $t$. We calculate the different fused noisy latents $z_t^{F1}$ with different $\lambda_t^{F1}$, then predicted the corresponding clean latents $\hat{z}_t^{F1}$ by the Eq. \ref{z0}. We decode each latents $\hat{z}_t^{F1}$ to a video $\{I^{F1}_i\}_{i=0}^N$ and calculate the average CLIP-IQA \cite{wang2023exploring} and Wrap Error (WE) \cite{lai2018learning} between all frames. The quality metrics CLIP-IQA and WE measure the visual quality and temporal temporal consistency of video, respectively. We rank all $M$+1 videos from 0 to $M$ (the lower the rank value, the better the quality) according to CLIP-IQA and WE, and obtain the rank values $R_{CLIP-IQA}$ and $R_{WE}$, respectively. We select the video with the lowest ($R_{CLIP-IQA}$+$R_{WE}$) as the video with best overall quality. And we select the corresponding $\lambda_t^{F1}$ value as the final $\lambda_t^{F1}$ at timestep $t$. Then set $\lambda_{t-1}^{c}$ to $\lambda_t^{F1}$ at timestep $t-1$ and repeat this best-of-$N$ selection for $\lambda_{t-1}^{F1}$. The range $r$, sample number $M$, and initial $\lambda_T^{F1}$ are hyper-parameters. 


In a similar way, we calculate the different fused noisy latents $z_t^{F2}$ with different $\lambda_t^{F1}$ and select the final $\lambda_t^{F1}$. After the confirmation of $\lambda_t^{F1}$ and $\lambda_t^{F2}$, we select the final $\lambda_t^{F}$ through fused noisy latents $z_t^{F}$.

\subsection{Temporal-strengthening Post-processing}

Fusing the latents between the IR/IE model and the T2V model can lead to better temporal consistency. However, the severe degree of temporal flickering of the IR model itself will still limit the performance. To solve this problem, we further propose to utilize the temporal prior in the I2V model Stable Video Diffusion (SVD) \cite{blattmann2023stable} for temporal-strengthening post-processing. First, we encode the video $\{I'_i\}_{i = 0}^N$ of IR/IE results (after latents fusion) into latents $z'_0$ with the VAE encoder of SVD. Since SVD is based on EDM sampling \cite{karras2022elucidating}, we invert $z'_0$ to noisy latents $z'_T$ with the inversion in the EDM framework:
\begin{align}
  z'_{t+1} &= \frac{\sigma_{t+1}z'_t + \left( \sigma_t - \sigma_{t + 1} \right) \cout^{t+1} \varepsilon'_{\theta} \left( \cin^{t} z'_{t}; \cnoise^{t+1} \right)}{\left( \sigma_t - \sigma_{t+1} \right) \left( 1 - \cskip^{t+1} \right) + \sigma_{t+1}} 
\end{align}
where $\sigma_t$, $\cskip^{t}$, $\cin^{t}$, $\cout^{t}$, and $\cnoise^{t}$ are parameters for noise scheduling in EDM framework, $\varepsilon'_{\theta}$ is the SVD denoising network. Then we apply EDM sampling from $z''_T$ ($z'_T$) to $z''_0$:
\begin{align}
  z''_t &= z''_{t+1} + \frac{\sigma_t - \sigma_{t+1}}{\sigma_{t+1}} \notag \\
        &\left( z''_{t+1} - \left( \cskip^{t+1} z''_{t+1} + \cout^{t+1} \varepsilon'_{\theta} \left( \cin^{t+1} z''_{t+1}; \cnoise^{t+1} \right) \right) \right)
\end{align}
When applying the SVD denoising network, the image condition is the first frame. Finally, we decode the latents $z''_0$ using the VAE decoder of SVD to obtain the final results $\{I''_i\}_{i = 0}^N$. Through the reconstrution of SVD, video $\{I''_i\}_{i = 0}^N$ has better temporal consistency than $\{I'_i\}_{i = 0}^N$.

\section{Experiments}

\subsection{Settings}

For homologous and heterogeneous T2V models, we use ZeroScope and CogVideoX-2B, respectively. For the I2V model in post-processing, we use Stable Video Diffusion. For all image and video diffusion models, the prompts are null texts. The $M$ in CoQ Fusion Ratio Strategy is set to 4. To accelerate the inference time, we apply the strategy every 10 timesteps and then apply the obtained $\lambda^{F1}$, $\lambda^{F2}$ and $\lambda^{F}$ to the following 9 timesteps. 

We evaluate our method on three video restoration tasks (zero-shot video super-resolution, zero-shot video deblurring and blind video super-resolution) and one video enhancement task (zero-shot low-light video enhancement). Following \cite{cao2025zero}, we collected 18 ground truth (GT) videos from REDS4, Vid4, and UDM10 for evaluation of zero-shot video super-resolution, collected 10 ground truth (GT) videos from the dataset REDS \cite{nah2019ntire} for evaluation of zero-shot video deblurring, and collected 10 paired low-normal videos from the DID dataset \cite{fu2023dancing} for evaluation of zero-shot low-light video enhancement. Due to the slowly sampling speed and a test video contains a lot of frames, we crop and resize the frames to 576$\times$320. For evaluation of blind super-resolution, we follow \cite{yeh2024diffir2vr,cao2025zero} and evaluate on DAVIS testing sets. For degraded videos of zero-shot video super-resolution, we follow the settings of linear degradation operator from \cite{rout2024solving} and \cite{fei2023generative}. For degraded videos of zero-shot video deblurring, we follow \cite{kwon2024solving,kwon2024vision} and use the temporal uniform blur kernel. For blind super-resolution, low-quality videos are generated using the real-world degradation pipeline of RealBasicVSR \cite{chan2022investigating}. For zero-shot video enhancement, we utilize the same degradation model in \cite{fei2023generative} and optimize the parameters of degradation model in the sampling process. The degradation model can be formulated as follows:
\begin{equation}
    y=f \mathcal{D}(\hat{z}_0)+\mathcal{M}
\label{eq:light}
\end{equation}
where the factor $f$ is a scalar and the mask $\mathcal{M}$ is a matrix of the same dimension as $\mathcal{D}(\hat{z}_0)$. We optimize $f$ and $\mathcal{M}$ using gradient descent during the sampling process.

Our method can be combined with the aggregation sampling in \cite{wang2024exploiting} to test on higher-resolution videos. For long video, we separate it into different video clips, neighbouring video clips share one overlapping frame. When our Temporal-Strengthening Post-Processing finishes the process of the previous video clip, the last/shared frame is used as the image condition in the next video clip to maintain the long-range temporal consistency.

All experiments were conducted on a 96G H20 GPU.

\begin{table*}[t]
\centering
\caption{Quantitative comparison with state-of-the-art methods for zero-shot 4$\times$ video super-resolution. The best results are highlighted in bold and the second best results are underlined. The WE and t-LPIPS values have been multiplied by 100.}
\setlength{\tabcolsep}{8pt} 
\resizebox{0.85\textwidth}{32mm}{
\addtolength{\tabcolsep}{-5pt}
\begin{tabular}{lcccccccccc}
\toprule
Methods              & Backbone  & PSNR$\uparrow$  & SSIM$\uparrow$  & CLIP-IQA$\uparrow$ & LPIPS$\downarrow$ & WE$\downarrow$ & FVD$\downarrow$ & DOVER$\uparrow$ & t-LPIPS$\downarrow$ & VMAF$\uparrow$\\
\hline
TDAN (sup.)	         & - &25.47	&0.7359	&0.3088	&0.1869	&0.5096	&155.3 &7.119 & 3.65 &79.99 \\
BasicVSR++ (sup.)	 & - &28.12	&0.8035	&0.4823	&0.1387	&0.4683	&122.5 &7.849 & 1.84 &82.75 \\
FMA-Net (sup.)	     & - &28.29	&\underline{0.8314}	&0.5121	&0.1357	&0.4923	&117.6 &8.207 & 1.06 & 84.56\\
VRT (sup.)	         & - & 29.58 & 0.8464 & 0.5224 &0.1221 &0.3264 & 93.1  & 8.928 & 0.66 & 84.08\\
MIA-SR (sup.)	     & - & \underline{29.63} & 0.8466 & 0.5320 & 0.1219 & 0.3315 & 94.6  &8.644 & 0.82 & 85.21\\
IART (sup.)	         & - & \textbf{29.69} & \textbf{0.8472} & 0.5329 & 0.1252 & 0.3212 & \underline{92.2} &9.105 &\underline{0.59} & 85.90 \\
\hline
PSLD	             & SDv1.5 &26.80	&0.7726	&0.4315	&0.1353	&0.8408	&140.4 &4.783  & 6.28 &75.80\\
PSLD+Text2Video	     & SDv1.5 &26.73	&0.7619	&0.4383	&0.1419	&0.8167	&139.1 &5.596  & 4.10 & 80.58 \\
PSLD+FateZero	     & SDv1.5 &26.81	&0.7852	&0.4186	&0.1475	&0.7522	&142.8 &6.332  & 3.59 & 82.24\\
PSLD+VidToMe	     & SDv1.5 &26.85	&0.7701	&0.4423	&0.1367	&0.6649	&138.3 &6.560  & 2.64 & 81.06\\
PSLD+FLDM	         & SDv1.5 &26.89	&0.7698	&0.4519	&0.1295	&0.6051	&130.2 & 6.937 & 2.37 & 82.63\\
PSLD+ZVRV	         & SDv1.5 &27.32	&0.7845	&\underline{0.5709}	&\underline{0.1110}	&\underline{0.2363}	&112.8 & \underline{9.240} & 0.62 &\underline{87.98} \\
\hline
PSLD                 & SDXL &27.51	&0.7802	&0.4485	&0.1361	&0.8523	&134.7 & 5.102 & 4.52 & 77.95\\
VISION-XL	         & SDXL &25.84	&0.7554	&0.2929	&0.2487	&0.4645	&333.3 & 6.781 & 3.79 & 76.33\\
PSLD+ZVRV	         & SDXL &28.35	&0.8099	&\textbf{0.5964} &\textbf{0.1072}	&\textbf{0.2220}	&\textbf{91.8} & \textbf{9.352} & \textbf{0.53} &\textbf{88.67} \\
\bottomrule
\end{tabular}
}
\label{ComparisonVSR}
\end{table*}

\begin{figure*}
    \centering
    \includegraphics[width=1.0\linewidth]{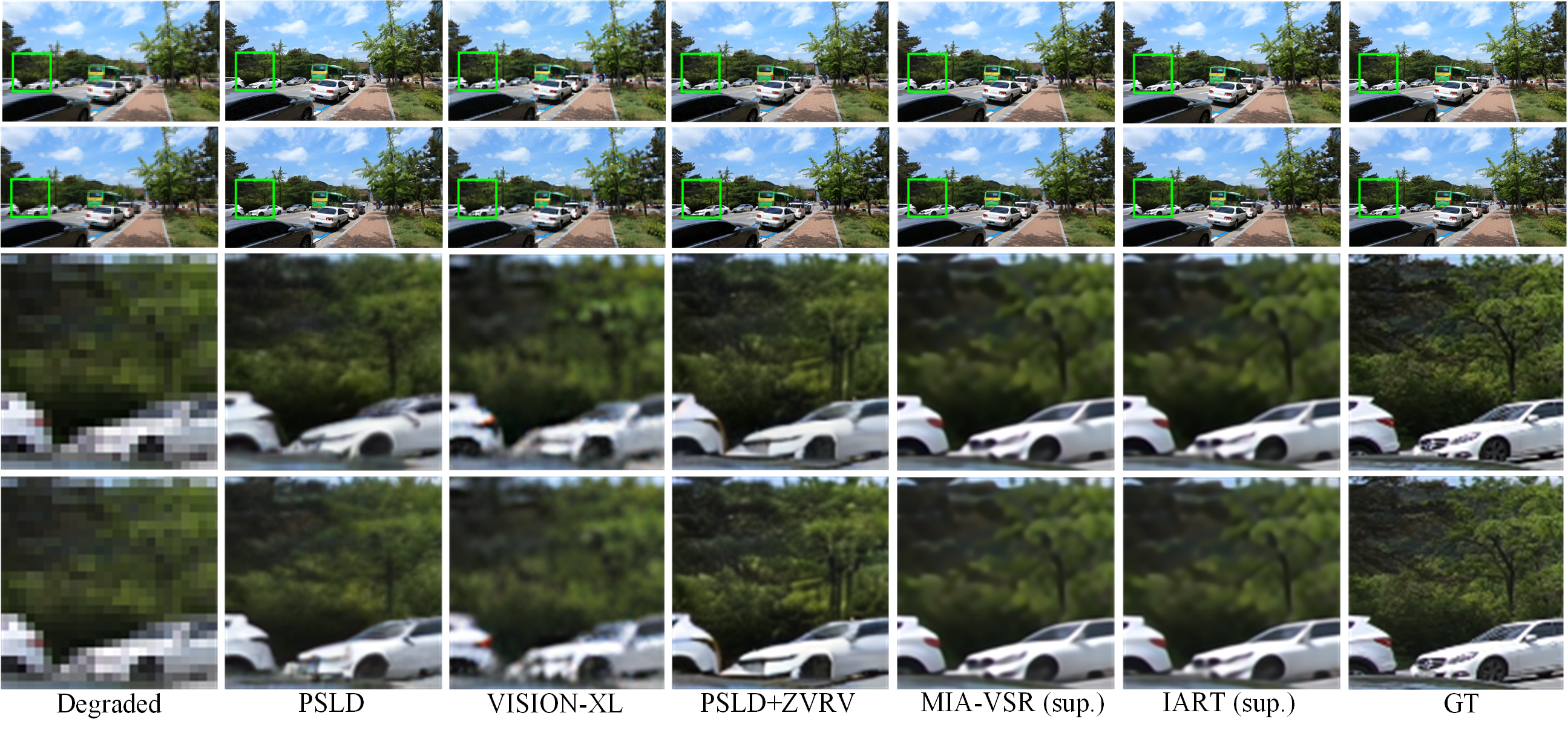}
    \caption{Visual quality comparison for zero-shot 4$\times$ video super-resolution. The first and second rows list the results of frames $t$ and $t+1$, the third and fourth rows list the results of the corresponding patches. Zoom in for better observation.}
    \label{fig:videosr}
\end{figure*}

\begin{table*}[t]
\centering
\caption{Quantitative comparison with state-of-the-art methods for zero-shot video deblurring. The best results are highlighted in bold and the second best results are underlined. The WE and t-LPIPS values have been multiplied by 100.}
\setlength{\tabcolsep}{8pt} 
\resizebox{0.8\textwidth}{13.5mm}{
\addtolength{\tabcolsep}{-5pt}
\begin{tabular}{lcccccccccc}
\toprule
Methods              & Backbone  & PSNR$\uparrow$  & SSIM$\uparrow$  & CLIP-IQA$\uparrow$ & LPIPS$\downarrow$ & WE$\downarrow$ & FVD$\downarrow$ & DOVER$\uparrow$ & t-LPIPS$\downarrow$ & VMAF$\uparrow$\\
\hline
VRT (sup.)	         & - &18.98 &0.4505 &0.2093 &0.7155 &\underline{0.1816} &1694.8 & \underline{7.581} & \underline{0.86} & \underline{81.94} \\
\hline
PSLD                 & SDXL &19.69	&0.4996	&0.1905	&0.4546	&1.5298	&662.0  & 5.573 & 3.12 & 76.02\\
SVI                  & -    &18.25	&0.4839	&0.2010	&0.5871	&1.3463	&1218.3 & 7.202 & 2.35 & 79.47 \\
VISION-XL	         & SDXL &\underline{19.82}	&\underline{0.5068}	&\underline{0.2147}	&\underline{0.4232}	&1.2623	&\underline{457.7} & 7.298 &1.84 & 80.66\\
PSLD+ZVRV	         & SDXL &\textbf{20.47}	&\textbf{0.5342} &\textbf{0.4813} &\textbf{0.3675}	&\textbf{0.1632} &\textbf{326.4} & \textbf{8.456} & \textbf{0.67} & \textbf{84.53} \\
\bottomrule
\end{tabular}
}
\label{ComparisonVDB}
\end{table*}

\begin{table*}[t]
\centering
\caption{Quantitative comparison with state-of-the-art methods for zero-shot low-light video enhancement. The best results are highlighted in bold and the second best results are underlined. The WE and t-LPIPS values have been multiplied by 100.}
\setlength{\tabcolsep}{8pt} 
\resizebox{0.83\textwidth}{13.5mm}{
\addtolength{\tabcolsep}{-5pt}
\begin{tabular}{lcccccccccc}
\toprule
Methods                                     & Backbone  & PSNR$\uparrow$  & SSIM$\uparrow$  & CLIP-IQA$\uparrow$ & LPIPS$\downarrow$ & WE$\downarrow$ & FVD$\downarrow$ & DOVER$\uparrow$ & t-LPIPS$\downarrow$ & VMAF$\uparrow$\\
\hline
FastLLVE (sup.)                             &-	     &10.62  &0.6911 &0.2524 & 0.2674	& 0.1299	 & 1465.2  & 2.093 & 15.76 & 49.33        \\
UDU (unsup.)                                &-	     &6.24  & 0.4346 &0.3560 & 0.3662   & \underline{0.1048} & 1618.7 &3.682 &10.39 & 60.01              \\
\hline
SGZ                                         &-	     &15.03	 &0.6987  &\underline{0.3835}	&0.1392	&0.2454	 &723.6 & \underline{5.715} & \underline{1.10} & \underline{81.81}           \\ 
PSLD                                        &SDv1.5  &\underline{20.37}	 &\underline{0.8281} &0.3086 &\underline{0.1186}  &0.6254  & \underline{547.3} & 5.102 & 9.34 & 65.63           \\
PSLD+ZVRV                                   &SDv1.5  &\textbf{20.69}  &\textbf{0.8453}	&\textbf{0.3955} &\textbf{0.0930} &\textbf{0.1017} & \textbf{251.9} & \textbf{8.965} &\textbf{0.78} &\textbf{83.22} \\
\bottomrule
\end{tabular}
}
\label{ComparisonLLVE}
\end{table*}

\begin{figure*}
    \centering
    \includegraphics[width=0.99\linewidth]{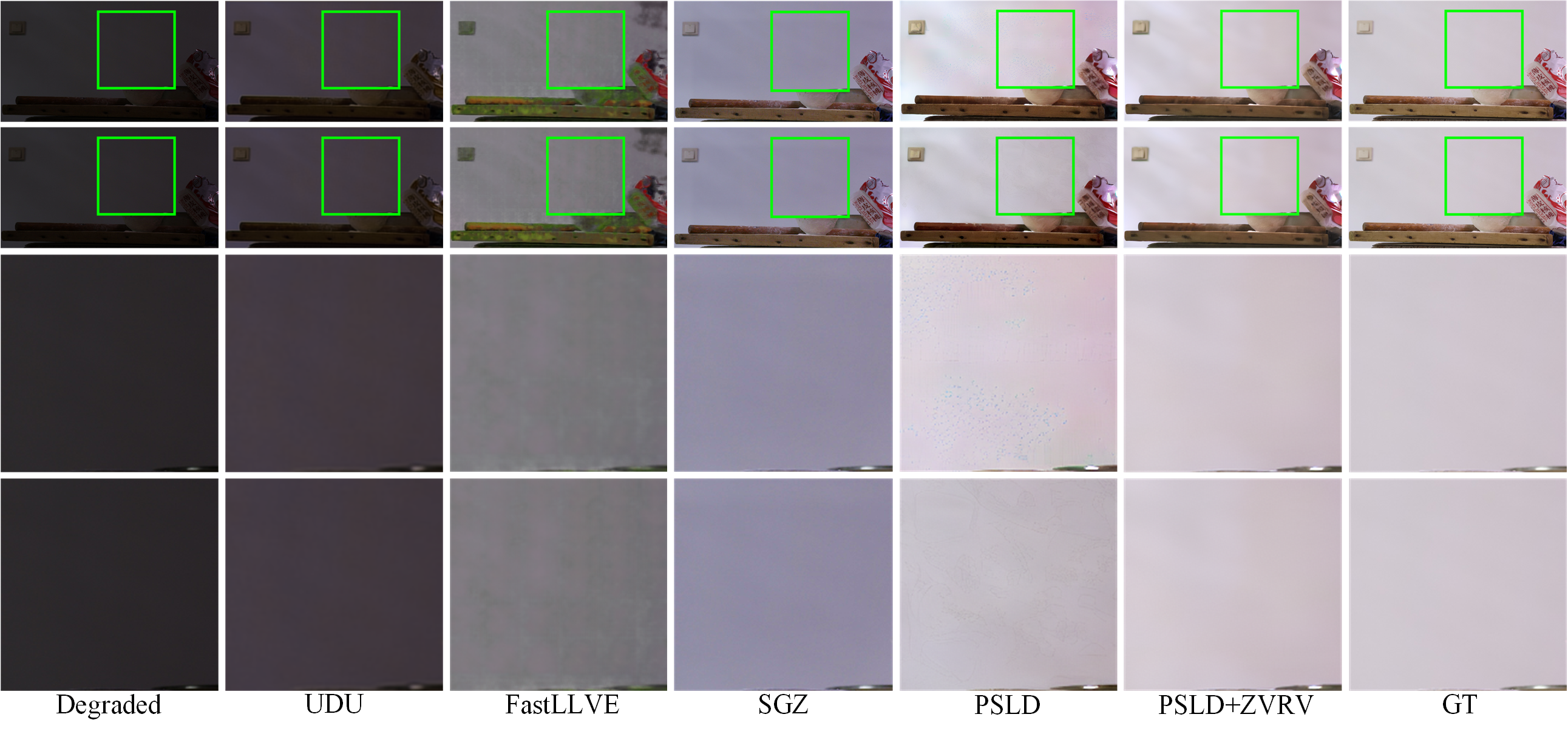}
    \caption{Visual quality comparison for zero-shot low-light video enhancement. The first and second rows list the results of frames $t$ and $t+1$, the third and fourth rows list the results of the corresponding patches. Zoom in for better observation.}
    \label{fig:videoenhance}
\end{figure*}

\subsection{Comparison with State-of-the-art Methods}

We utilize nine metrics to evaluate the restoration and enhancement quality. Besides the commonly used metrics PSNR, SSIM, we utilize CLIP-IQA and LPIPS to evaluate the visual perceptual quality, utilize Warping Error (WE) \cite{lai2018learning}, FVD \cite{unterthiner2019fvd}, DOVER \cite{wu2023exploring}, t-LPIPS \cite{perez2018perceptual,chu2020learning}, and VMAF \cite{rassool2017vmaf,zheng2025efficient} to evaluate the temporal consistency. For zero-shot video super-resolution, we compare our method with supervised (sup.) methods TDAN \cite{tian2020tdan}, BasicVSR++ \cite{chan2022basicvsr++}, FMA-Net \cite{youk2024fma}, VRT \cite{liang2024vrt}, MIA-VSR \cite{zhou2024video}, IART \cite{xu2024enhancing}, and zero-shot video restoration method VISION-XL \cite{kwon2024vision}. We also adding zero-shot video editing methods Text2Video-Zero \cite{khachatryan2023text2video}, FateZero \cite{qi2023fatezero}, VidToMe \cite{li2024vidtome}, FLDM \cite{lu2024fuse} for comparison. For these zero-shot video editing methods, we transfer them to backbone PSLD. For zero-shot video deblurring, we compare our method with supervised (sup.) method VRT \cite{liang2024vrt} and zero-shot video restoration method SVI \cite{kwon2024solving} and VISION-XL \cite{kwon2024vision}. Since SVI only releases the code for zero-shot video deblurring, we do not compare it on other restoration tasks. For zero-shot low-light video enhancement, we compare our method with supervised (sup.) method FastLLVE \cite{li2023fastllve}, unsupervised (unsup.) method UDU \cite{zhu2024unrolled} and zero-shot video enhancement method SGZ \cite{zheng2022semantic}. For blind video super-resolution, we compare our method with image super-resolution methods (DiffBIR \cite{lin2024diffbir}, DiT4SR \cite{duan2025dit4sr}, TSD-SR \cite{dong2025tsd}) and video super-resolution methods (Upscale-A-Video \cite{zhou2024upscale}, SeedVR-7B \cite{wang2025seedvr}, DiffIR2VR \cite{yeh2024diffir2vr}, and ZVRD \cite{cao2025zero}). DiffBIR, DiT4SR, and TSD-SR are also used as our backbones.

\vspace{-0.21cm}
For a fair comparison, we change the stable diffusion backbone from Stable Diffusion v1.5 (SDv1.5) to Stable Diffusion XL (SDXL) when comparing with VISION-XL, which also uses SDXL as its backbone.  Since the T2V model Zeroscope shares the same VAE encoder with SDv1.5, but the VAE encoder of SDXL is fine-tuned from that of SDv1.5, Zeroscope and SDXL do not use the same latent space. Thus, homologous latent fusion cannot be applied, we only apply the residual modules. Since DiT4SR is based on flow-matching sampling, which is inconsistent with the DDIM sampling used in T2V models ZeroScope and CogVideoX-2B, neither homologous nor heterogeneous latent fusion can be applied. Therefore, we only apply temporal-strengthening post-processing. For the one-step diffusion model TSD-SR, latent fusion is also inapplicable, only temporal-strengthening post-processing is employed. This also demonstrates that our method can be applied to any image-based model, regardless of its latent space or sampling method/steps.



\begin{table*}[t]
\centering
\caption{Quantitative comparison with state-of-the-art methods for 4$\times$ blind video super-resolution on the DAVIS dataset. The best results are highlighted in bold and the second best results are underlined. The WE and t-LPIPS values have been multiplied by 100.}
\setlength{\tabcolsep}{8pt} 
\resizebox{0.825\textwidth}{23mm}{
\addtolength{\tabcolsep}{-5pt}
\begin{tabular}{lcccccccccc}
\toprule
Methods                                  & Backbone  & PSNR$\uparrow$  & SSIM$\uparrow$  & CLIP-IQA$\uparrow$ & LPIPS$\downarrow$ & WE$\downarrow$ & FVD$\downarrow$ & DOVER$\uparrow$ & t-LPIPS$\downarrow$ & VMAF$\uparrow$\\
\hline
Upscale-A-Video                          &SDv2.0 &23.76 & 0.5676 &0.7028 &0.3114 & 0.5958 & 585.3 & 5.962 & 1.06 &79.49         \\
SeedVR                                   &SDv3.0 &22.27 & 0.5162 &0.8268 &0.3070 & 0.5634 & 717.2 & 8.739 & 1.23 &83.61          \\
\hline
DiT4SR                                   &SDv3.0 &23.41 & 0.6235 &\underline{0.9283} &0.2944 & 1.8274 & 917.6 & 4.305 &11.29 & 73.24          \\
DiT4SR+ZVRV                              &SDv3.0 &23.66 & 0.6387 &\textbf{0.9363} &0.2909 & 0.5526 & 839.7 & 7.821 &1.20 & 82.80                                        \\
\hline
TSD-SR                                   &SDv3.0 &25.15 & 0.6809 &0.8702 &0.1985 & 1.0295 & 357.8 & 6.464 &2.12 & 77.93         \\
TSD-SR+ZVRV                              &SDv3.0 &25.22 & 0.6866 &0.8761 &0.1811 & \underline{0.4731} & 329.2 & \underline{8.983} &0.64 &\underline{84.36} \\
\hline
DiffBIR                                  &SDv2.1 &26.50	 &0.6869	&0.8340	  &0.1751 &0.8061 &279.1 & 5.720 &3.92 & 76.64        \\
DiffIR2VR                                &SDv2.1 &26.63	 &0.6904	&0.8152	  &0.1743 &0.7504 &273.9 & 6.369 &3.63 & 80.25	              \\
DiffBIR+ZVRD                             &SDv2.1 &\underline{26.86}	 &\underline{0.7029}	&0.8017	  &\underline{0.1565} &0.5042 &\underline{262.5} & 7.551 & \underline{0.55} & 81.02 \\
DiffBIR+ZVRV                             &SDv2.1 &\textbf{27.42}  &\textbf{0.7388}	&0.8691 &\textbf{0.1395}  &\textbf{0.3755}	 &\textbf{231.7} & \textbf{9.076} & \textbf{0.41} & \textbf{86.37}         \\
\bottomrule
\end{tabular}
}
\label{ComparisonBVSR}
\end{table*}

\begin{figure*}
    \centering
    \includegraphics[width=1.0\linewidth]{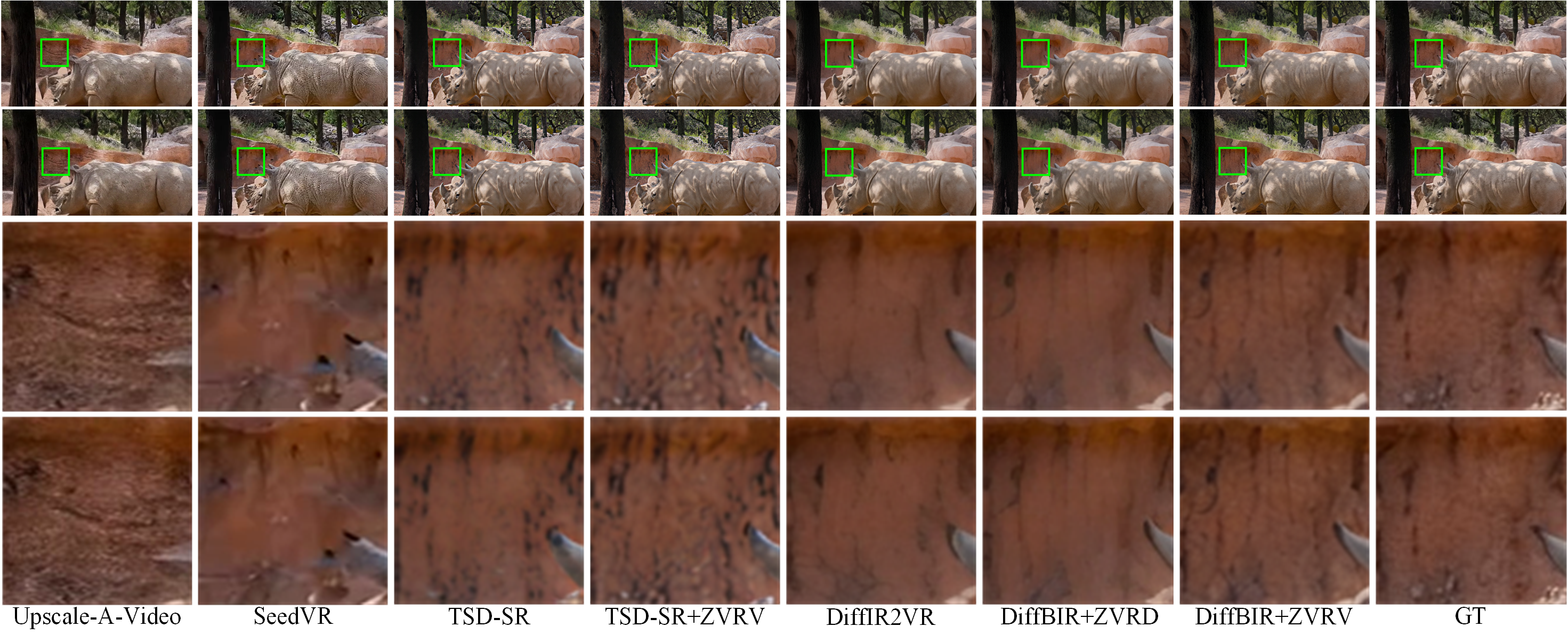}
    \caption{Visual quality comparison for 4$\times$ blind video super-resolution. The first and second rows list the results of frames $t$ and $t+1$, the third and fourth rows list the results of the corresponding patches. Zoom in for better observation.}
    \label{fig:videobsr}
\end{figure*}

Table \ref{ComparisonVSR}, \ref{ComparisonVDB}, \ref{ComparisonLLVE}, and \ref{ComparisonBVSR} list the quantitative results on the evaluation data for zero-shot video super-resolution, zero-shot video deblurring, zero-shot low-light video enhancement and blind video super-resolution, respectively. It can be observed that our method ZVRV outperforms PSLD on nine metrics for both zero-shot tasks. For zero-shot 4$\times$ video super-resolution, our method outperforms PSLD (with SDXL backbone) with 0.84 dB gain for PSNR, 0.1479 gain for CLIP-IQA, 4.25 gain for DOVER, 3.99 gain for t-LPIPS, 10.72 gain for VMAF and only has nearly 1/4 WE value. Our method also outperforms the supervised method IART on seven metrics. For zero-shot video deblurring, it can be observed that our method outperforms the compared SOTA methods on all nine metrics. For zero-shot low-light video enhancement, our method outperforms PSLD on nine metrics and only has nearly 1/6 WE value. Our model can also be applied to blind video restoration by being inserted into blind image restoration in a zero-shot manner. For blind video super-resolution, our method with DiffBIR backbone achieves the best performance on eight metrics except CLIP-IQA, and our method with DiT4SR achieves the best performance on CLIP-IQA. Our method outperforms the SOTA method ZVRD with 0.56 dB gain for PSNR, 0.0359 gain for SSIM, 0.0674 gain for CLIP-IQA, 1.525 gain for DOVER, 5.35 gain for VMAF. In addition, our method, which utilizes the DiffBIR/TSD-SR backbone, significantly outperforms the SOTA methods Upscale-A-Video and SeedVR across all nine metrics. Table \ref{ComparisonSpeed} lists the inference time of comparison method per frame at a resolution of 576$\times$320. It can be observed that our method with the TSD-SR backbone has the fastest inference time.

Figs. \ref{fig:videosr}, \ref{fig:videoenhance}, and \ref{fig:videobsr} present the visual comparison results on the evaluation data for zero-shot video super-resolution, zero-shot low-light video enhancement and blind video super-resolution, respectively. For video super-resolution, the areas of tree and car have different texture on the two frames of PSLD results. The results of VISION-XL, MIA-VSR, and IART are blurry. Our method can restore more sharp and temporal consistent textures. For low-light video enhancement, UDU, FastLLVE and SGZ have severe color shifts. There are color shift, different texture and brightness between two frames of PSLD. Our method preserve better temporal consistency. For blind video super-resolution, our method can also restore more temporally consistent textures on the rock, which is also closest to the ground truth. The result of Upscale-A-Video has artifacts, the result of SeedVR is blurry, and neither is faithful to the ground truth.

\vspace{-1.2pt}
We also conducted a user study to further evaluate the visual quality and temporal consistency. To facilitate the comparison, we only compared the zero-shot video super-resolution results of PSLD, VISION-XL, and PSLD+ZVRV on eight groups of videos. Each user is asked to evaluate the visual quality and temporal consistency of the video with a score ranged from 1 to 3, where 3 indicates good
quality and 1 indicates bad quality. The average score for PSLD, VISION-XL, and PSLD+ZVRV is 2.07, 1.05 and 2.91, respectively, which demonstrates that our method has much better visual quality and temporal consistency. We also provide a demo video. 

\begin{table}[t]
\centering
\caption{Inference Time on GPU 96G H20.}
\setlength{\tabcolsep}{2pt}
\resizebox{0.5\textwidth}{5mm}{
\begin{tabular}{lccccc}
\toprule
Methods           & Upscale-A-Video  & SeedVR & DiffBIR+ZVRV & DiT4SR+ZVRV  & TSD-SR+ZVRV  \\
\hline
Time (s) $\downarrow$  &4.12  &10.55  & 61.29  &21.83 &2.96\\
\bottomrule
\end{tabular}
}
\label{ComparisonSpeed}
\vspace{-0.25cm}
\end{table}

\begin{table*}
\centering
\caption{Ablation study on the CoQ fusion ratio strategy (with different hyperparameters $M$ and $r$) and the fixed fusion ratio strategy. The best results are highlighted in bold.}
\setlength{\tabcolsep}{4pt}
\resizebox{0.875\textwidth}{14mm}{
\begin{tabular}{ccccccccccc}
\toprule
\multicolumn{2}{l}{Fusion Ratio Strategy}                  & PSNR$\uparrow$  & SSIM$\uparrow$  & CLIP-IQA$\uparrow$ & LPIPS$\downarrow$ & WE$\downarrow$ & FVD$\downarrow$ & DOVER$\uparrow$ & t-LPIPS$\downarrow$ & VMAF$\uparrow$\\
\hline
\multirow{5}{*}{CoQ Fusion Ratio Strategy} &$M$=2, $r$=0.45  &26.74  &0.6953	&0.8472 &0.1649  &0.4186  &258.5 & 7.251 & 0.92 & 83.42    \\\cline{2-11}
                                                 &$M$=3, $r$=0.45  &27.15  &0.7196	&0.8599 &0.1573  &0.3952  &240.8 & 8.854 & 0.63 & 85.01    \\\cline{2-11}
                                                 &$M$=4, $r$=0.45  &\textbf{27.42}  &\textbf{0.7388} &\textbf{0.8691} &\textbf{0.1395}  &\textbf{0.3755}  &\textbf{231.7} & \textbf{9.076} & \textbf{0.41} & \textbf{86.37}    \\\cline{2-11}
                                                 &$M$=4, $r$=0.50  &27.20  &0.7316	&0.8610 &0.1432  &0.3764  &239.5 & 8.983 & 0.51 & 86.04    \\\cline{2-11}
                                                 &$M$=4, $r$=0.40  &27.09  &0.7272	&0.8623 &0.1408  &0.3783  &234.1 & 8.779 & 0.50 & 85.79    \\\hline
\multicolumn{2}{l}{Fixed Fusion Ratio Strategy}                &26.59  &0.6915	&0.8347 &0.1724  &0.4298  &265.2 & 6.451 & 0.98 & 82.26\\
\bottomrule
\end{tabular}
}
\label{AblationFusionRatio}
\end{table*}

\subsection{Ablation Study}

\begin{table}
\centering
\caption{Ablation study for Homologous Latents Fusion (HMLF), Heterogenous Latents Fusion (HTLF), CoQ Fusion Ratio Strategy (CFRS) and Temporal-Strengthening Post-Processing (TSPP) on 4$\times$ blind video super-resolution task. The WE and t-LPIPS values have been multiplied by 100.}
\resizebox{0.5\textwidth}{30.5mm}{
\begin{tabular}{ccccccc}
\toprule
HMLF              & $\times$   & $\checkmark$ & $\checkmark$     & $\checkmark$  & $\checkmark$      \\\hline                          
HTLF              & $\times$   & $\times$     & $\checkmark$     & $\checkmark$  & $\checkmark$     \\\hline
CFRS              & $\times$   & $\times$     & $\times$         & $\checkmark$      & $\checkmark$  \\\hline
TSPP             & $\times$   & $\times$     & $\times$         & $\times$  & $\checkmark$      \\\hline
PSNR$\uparrow$                  & 26.50      & 26.69	    & 27.12	           & 27.35        &27.42     \\
SSIM$\uparrow$                  & 0.6869     & 0.6981       & 0.7201           & 0.7256       &0.7388    \\
CLIP-IQA$\uparrow$              & 0.8340     & 0.7339       & 0.7026           & 0.8446       &0.8691    \\
LPIPS$\downarrow$               & 0.1751     & 0.1702       & 0.1678           & 0.1425       &0.1395    \\
WE$\downarrow$                  & 0.8061     & 0.6957       & 0.5536           & 0.4942       &0.3755    \\
FVD$\downarrow$                 & 279.1      & 273.8        & 255.6            & 248.2        &231.7    \\
DOVER$\uparrow$                 &5.720       & 6.005        & 7.028            & 7.459        &9.076  \\
t-LPIPS$\downarrow$             &3.92        & 3.36         & 1.98             & 1.63         &0.41  \\
VMAF$\uparrow$                  &76.64       & 78.35        & 80.96            & 82.20        &86.37\\
Time(s)$\downarrow$             &9.15        & 22.72        & 51.98            & 58.69        &61.29         \\
Memory(GB)$\downarrow$          &8.3         & 16.0         & 54.2             & 54.2         &72.8         \\
Params(B)$\downarrow$           &1.4         & 3.1          & 5.1              & 5.1          &7.4         \\
\bottomrule
\end{tabular}
}
\label{Ablation}
\end{table}

In this section, we perform an ablation study to demonstrate the effectiveness of the proposed Homologous Latents Fusion, Heterogenous Latents Fusion, CoQ Fusion Ratio Strategy, and Temporal-Strengthening Post-Processing. Besides the nine metrics, we also list the runtime per frame at a resolution of 576×320, the peak GPU memory usage, and the number of parameters when each module is added.

Taking 4$\times$ blind video super-resolution as an example, Table \ref{Ablation} lists the quantitative comparison results in the evaluation data by adding these modules one by one. When the CoQ Fusion Ratio Strategy is not added, Homologous and Heterogenous Latents Fusion use a simple linear update schedule. It can be observed that both Homologous and Heterogenous latents can significantly improve the temporal consistency. Homologous Latents Fusion can bring 0.19 dB gain for PSNR, 0.0112 gain for SSIM, and 0.1104 gain for WE. Heterogenous Latents Fusion can bring 0.43 dB gain for PSNR, 0.022 gain for SSIM, 0.1421 gain for WE, and 18.2 gain for FVD. However, Homologous Latents Fusion results in an increase of 13.57 seconds in runtime, 7.7 GB in peak GPU memory, and 1.7B in parameters, Heterogenous Latents Fusion results in an increase of 29.26 seconds in runtime, 38.2 GB in peak GPU memory, and 2B in parameters. The CoQ Fusion Ratio Strategy can significantly improve the value of CLIP-IQA and LPIPS. It also brings 0.23 dB gain for PSNR, and 0.0594 gain for WE. And it results in an increase of 6.71 seconds in runtime. Temporal-Strengthening Post-Processing can further reduce the WE and FVD value. It only adds 2.65 seconds in runtime, but results in an increase of 18.6 GB in peak GPU memory usage and 2.3B in parameters. Our method with DiffBIR backbone, has 7.4B parameters but achieves significantly better performance than SeedVR-7B, which has 7B parameters.

We also conduct an ablation study on the CoQ fusion ratio strategy (with different hyperparameters \( M \) and \( r \)) and the fixed fusion ratio strategy. For the fixed fusion ratio strategy, we set \( \lambda^{F1} \), \( \lambda^{F2} \), and \( \lambda^{F} \) to 0.1, 0.01, and 0.5, respectively, for all timesteps. For hyperparameter \( r \), the value is halved after each search. Taking 4$\times$ blind video super-resolution as an example, Table \ref{AblationFusionRatio} lists the quantitative comparison results. It can be observed that the CoQ fusion ratio strategy is more sensitive to the hyperparameter \( M \). As \( M \) increases, the performance improves as well. Considering the trade-off between speed and performance, we set \( M = 4 \) and \( r = 0.45 \) , as this combination achieves the best performance. Besides, we apply the CoQ fusion ratio every 10 timesteps and apply the obtained $\lambda^{F1}$, $\lambda^{F2}$, and $\lambda^{F}$ to the following 9 timesteps. Although applying the CoQ fusion ratio more frequently can result in better performance, we choose to apply it every 10 timesteps considering the trade-off between speed and performance. When \( M \) ranges from 2 to 4, and \( r \) ranges from 0.4 to 0.5, the CoQ fusion ratio strategy consistently outperforms the fixed fusion ratio strategy.

\section{Conclusion}

In this paper, we propose  zero-shot video restoration and enhancement with assistance of video diffusion models. Through the proposed homologous and heterogenous latents fusion, we can utilize any kind of SOTA T2V model to assist the image restoration/enhancement model in achieving temporal consistent video restoration/enhancement. We further propose the CoQ fusion ratio strategy to better control the fusion ratio when fusing latents at each timestep. Our temporal-strengthening post-processing further overcomes the limitations of the IR model's sampling method and steps, enabling it to be applied to any IR model. Experimental results demonstrate the superiority of the proposed method in performance and temporal consistency.

%

%

\ifCLASSOPTIONcaptionsoff
  \newpage
\fi



\bibliographystyle{IEEEtran}
\bibliography{egbib}
\end{document}